\documentclass[conference]{ieeeconf}

\IEEEoverridecommandlockouts                              

\usepackage{amsmath,amssymb,amsfonts}
\usepackage[english]{babel}
\usepackage{graphicx}
\usepackage{textcomp}
\usepackage{xcolor}
\usepackage{url}
\usepackage{regexpatch}
\usepackage[colorinlistoftodos,prependcaption]{todonotes}

\usepackage[absolute,overlay]{textpos}

\usepackage{subcaption}

\def\BibTeX{{\rm B\kern-.05em{\sc i\kern-.025em b}\kern-.08em
		T\kern-.1667em\lower.7ex\hbox{E}\kern-.125em}}

\makeatletter
\newcommand{\linebreakand}{%
\end{@IEEEauthorhalign}
\hfill\mbox{}\par
\mbox{}\hfill\begin{@IEEEauthorhalign}
}
\makeatother

\makeatletter
\xpatchcmd{\@todo}{\setkeys{todonotes}{#1}}{\setkeys{todonotes}{#1}}{}{}

\usepackage{csquotes}
\usepackage{tcolorbox}
\usepackage[hidelinks]{hyperref}
\usepackage{dblfloatfix}
\makeatother

\begin{document}

	\title{\LARGE \bf
The Emotional Dilemma: Influence of a Human-like Robot on Trust and Cooperation
	}




\author{Dennis Becker$^{*}$, Diana Rueda$^{*}$, Felix Beese, Brenda Scarleth Gutierrez Torres$^{\dag}$, Myriem Lafdili,\\ Kyra Ahrens, Di Fu, Erik Strahl, Tom Weber, Stefan Wermter
\thanks{$^{*}$These authors contributed equally. }%
\thanks{$^{\dag}$Stipendiary from CONACyT and DAAD }%
\thanks{The authors gratefully acknowledge partial support from the German Research Foundation DFG under project CML, TRR169 and LeCareBot.}
\thanks{All of the authors are with the Knowledge Technology Group, Department of Informatics, Universität Hamburg, Vogt-Kölln-Straße 30, Hamburg D-22527, Germany.}%
\thanks{Many thanks to Moritz Lahann, Ramtin Nouri, Jose Angel Sanchez Castro, Sebastian Stelter, and Denyse Uwase.}%
}
      
	
	\maketitle
	\begin{abstract}
%
Increasing anthropomorphic robot behavioral design could affect trust and cooperation positively. However, studies have shown contradicting results and suggest a task-dependent relationship between robots that display emotions and trust.
Therefore, this study analyzes the effect of robots that display human-like emotions on trust, cooperation, and participants' emotions. In the between-group study, participants play the coin entrustment game with an emotional and a non-emotional robot.
The results show that the robot that displays emotions induces more anxiety than the neutral robot. Accordingly, the participants trust the emotional robot less and are less likely to cooperate. Furthermore, the perceived intelligence of a robot increases trust, while a desire to out-compete the robot can reduce trust and cooperation.
Thus, the design of robots expressing emotions should be task dependent to avoid adverse effects that reduce trust and cooperation.

%
%
%
	\end{abstract}
	

\begin{textblock*}{16cm}(2cm,1cm) 
\begin{minipage}{16cm}
In: 32nd IEEE International Conference on Robot and Human Interactive Communication (RO-MAN), Busan, Korea, 28 Aug - 31 Aug, 2023.
\end{minipage}
\end{textblock*}

\begin{textblock*}{16cm}(2.5cm,26.1cm)
\fbox{\begin{minipage}{16cm}
© 2023 IEEE. Personal use of this material is permitted. Permission from IEEE must be obtained for all other uses, in any current or future media, including reprinting/republishing this material for advertising or promotional purposes, creating new collective works, for resale or redistribution to servers or lists, or reuse of any copyrighted component of this work in other works.
\end{minipage}}

\end{textblock*}

\section{Introduction}
\label{sec:introduction}


Trust and cooperation affect humans' willingness to interact with robots~\cite{Freedy2007}, and are essential to establish successful human-robot interaction~\cite{Yagoda2012,Khavas2021,Lewis2018}.
Human-robot trust is composed of human-related, robot-related, and environment-related factors~\cite{Hancock2011,Mittu2016}.
To the human-related factors also belong emotions, which can influence trust~\cite{Lee2012} and have been suggested as an integral part of forming trust~\cite{Miller2021}. Humans' emotional states have been shown to affect trust in human-robot interaction~\cite{Stock-Homburg2022}, where positive emotions such as happiness increase trust~\cite{Stokes2010} and negative emotions such as anxiety reduce trust~\cite{Kraus2020}.
Likewise, trust in a robot influences the willingness to cooperate with the robot~\cite{Jones1998}, which can be more pronounced in cooperative tasks~\cite{Weidemann2021}. Cooperation is frequently associated with an initial investment and trust in the collaborator in expectation of a greater mutual benefit~\cite{Rand2013}.

While robots are becoming steadily human-like, research has been conducted on how anthropomorphic a robot should be designed for a specific task~\cite{Haring2013,Desai2009}. Studies suggest that robots that act more human-like by expressing emotions are perceived as more trustworthy~\cite{Martelaro2016,Hashemian2019} and encourage cooperation~\cite{Takahashi2021,Andriella2021}. Specifically, trust is fostered when the robot's expressed emotions are congruent and engages in social conversation~\cite{Paradeda2016}. Contrary, to this relationship of increased anthropomorphism and trust, recent findings suggest a task dependency~\cite{Onnasch2022,Roesler2021}, where a robot expressing social cues for critical tasks receives less trust than a neutral robot~\cite{Rossi2020}.
Thus, research on human-like robots and their effect on trust and cooperation with respect to emotions is required.

For measuring trust and cooperation in human-robot interaction, we adopt a variation of the prisoner's dilemma, the coin entrustment game~\cite{Yamagishi2005}. The coin entrustment game allows participants to decide on the amount of trust in the robot in expectation of an increased reward. In the conducted between-group experiment, an emotional robot is compared to a non-emotional robot.
While the non-emotional robot portrays neutral emotions, the emotional robot displays happiness or sadness, depending on the participant's cooperation. The emotions are conveyed utilizing multiple modalities, such as speech interjections that deploy emotional prosody \cite{Savery2021}, body gestures~\cite{Saunderson2019}, and facial expressions~\cite{Rawal2022}.

\begin{figure}[t]
\centering
    \includegraphics[width=0.7\linewidth]{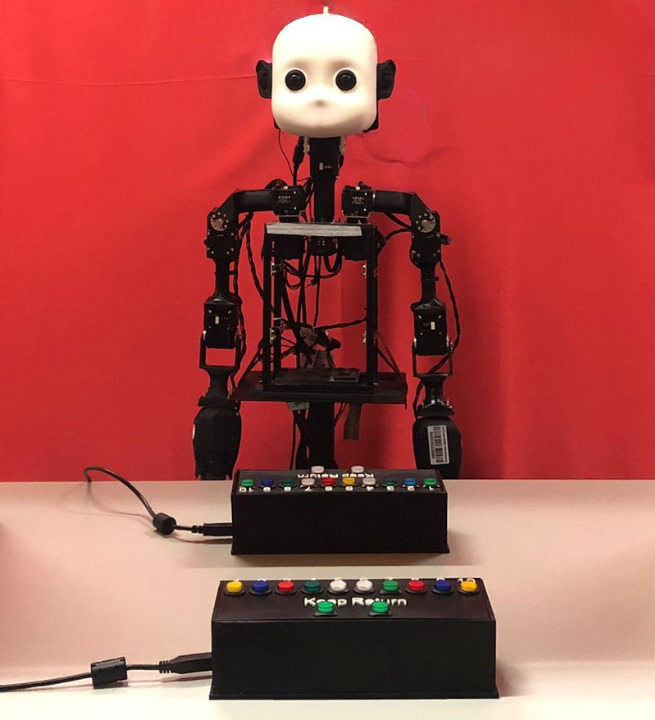}
    \caption{Coin entrustment game to measure the difference in trust and cooperation with an emotional and non-emotional robot.}
    \label{fig:nicos_stand}
\end{figure}



\section{Related Work}
\label{sec:related_work}

The prisoner's dilemma is a game theory-based~\cite{Axelrod84} social dilemma that has frequently been used to study trust and cooperation in social situations~\cite{VanLange2013}.
The initial definition accounts for one round, where defecting is the preferred strategy to avoid punishment. However, experiments could not reliably explain the emergence of participants cooperating~\cite{Grandgirard2002}. Thus, the repeated prisoner's dilemma~\cite{rapoport1965prisoner} was suggested, where the game is played over consecutive rounds. With an undisclosed number of rounds, a cooperative strategy is encouraged and leads to a superior outcome for both players~\cite{Garcia2018}.
Since the prisoner's dilemma has a predefined payout matrix, a coin entrustment variant of the game was proposed~\cite{Yamagishi2005}. In the coin entrustment game, a player's pay off and risk are dependent on the amount of entrusted coins, which allows measuring trust in the other player to cooperate and own cooperation separately.


Research on social dilemmas suggests that humans act more selfishly and less emotional when interacting with artificial agents and robots~\cite{Chugunova2022,March2021,Hoegen2015}. When facing an artificial agent, humans tend to derive decisions more rational~\cite{Houser2016}, and the influence of emotions on decision-making is mitigated~\cite{Schniter2020}. These findings are supported by physiological measures of skin conductivity and heart rate in a betting scenario against a computerized agent and a human~\cite{Teubner2015}.
Studies on the prisoner's dilemma in human-robot interaction suggest that cooperative behavior with humans tends to be higher than with robots~\cite{Sandoval2016,Maggioni2022}.
When cooperating with a robot, self-reported measures indicate that decision-making is less emotionally driven and imply a notable lack of empathy that could result in a more selfish behavior~\cite{RePEc:jau:wpaper:2021/10}.
However, when the robot is facing consequences for losing in the prisoner's dilemma, such as erasing its memory, individuals' game-play becomes more empathetic towards the robot~\cite{Hsieh2020}. 
Similarly, a robot expressing moral values leads to less competitive game-play, while a robot expressing emotions can increase competitive behaviour~\cite{PLAKS2022107301}.
Contradicting these cooperative preferences, research on the n-player prisoner's dilemma with human and robotic players shows that cooperation with robotic players is higher, potentially due to the unpredictability of human players' strategy~\cite{Duffy2016}. Correspondingly, for the coin entrustment game, more trust towards robotic players is reported~\cite{wu2016trust}.


Non-verbal communication signals and indicates willingness to cooperate in social dilemma and trust games~\cite{Jahng2017,Mou2020}. 
People who are likely to cooperate are emotionally more expressive than people that are inclined to defect~\cite{Schug2010}.
Additionally, research has shown that humans are sensitive to emotional display and facial expressions when evaluating a person's cooperativeness~\cite{Krumhuber2007}. Specifically, smiling can evoke and indicate cooperation and is perceived as an indicator of the other's trustworthiness~\cite{Brown2002}, whereas contemptuous behavior is associated with defect~\cite{Reed2012}. 
Likewise, body motions~\cite{Strassmann2016} and gestures~\cite{Zorner2021} of non-verbal communication are utilized to judge the partner's tendency to cooperate.
Moreover, vocal features affect trustworthiness and cooperation~\cite{Torre2021}, where emotional expressiveness~\cite{schug2010emotional} and a happy-sounding voice can foster trust~\cite{torre2020if}. Further, congruence between a robot's behavior and its voice influences trust~\cite{Torre2018} and convergence of a robot's speech influences cooperation~\cite{Manson2013}.
Despite the effect of non-verbal communication, studies on the prisoner's dilemma with a virtual agent suggest, that nonverbal behavior might be too subtle to be recognized~\cite{Strassmann2018}. Subsequently, no difference in the cooperation between a robot expressing sad and angry emotions could be shown; however, recognizing the robot's emotions can reduce the participants' cooperation~\cite{Hsieh2022}.



\section{Methodology}
\label{sec:methodology}

\subsection{Participants}
\label{subsec:participants}

After receiving a positive response from the Ethics Commission of the Department of Informatics at the University of Hamburg,
participants were recruited from the university's campus, and the experiment was conducted with 47 participants. 
Out of these participants, three were excluded due to technical issues, and an additional three participants were discarded based on the control question. As control question, the participants rated the robot's emotionality on a 5-point scale ranging from Emotional to Non-Emotional. Participants who considered the non-emotional robot as Emotional, and participants that labeled the emotional robot as Non-Emotional were excluded. This results in a data set of 41 participants for analysis, consisting of 20 participants in the emotional and 21 participants in the non-emotional group. 

Of these participants, 41.5\% were female and 58.5\% were male. The participants' age distribution was 78\% between 18--29 years, 19.5\% between 30--39 years, and 2.5\% between 40--49 years. 
Four of the participants self-reported a high familiarity with humanoid robots, two participants stated a high familiarity with negotiation games, and one participant self-reported a high familiarity with both humanoid robots and negotiation games. 


\subsection{Experiment Design}
\label{subsec:experiment_design}


The experiment implements the coin entrustment game, and the participants play a total of 16 rounds, where each round consists of two stages. During the first stage, both the participant and the robot secretly entrust between one and ten coins to the other player.
For the robot, the number of entrusted coins follows the design of the experiment conducted in~\cite{wu2016trust}. Specifically, the number of entrusted coins depends on the payoff from the previous round and at least one coin is entrusted. The number of entrusted coins is expressed as follows:



$$
E(p) = \left\{\begin{array}{ll}
    \lceil min(10 + \frac{p-10}{1.5}, 10) \rceil & \text{if } p > 0, \\
    1 & \text{if } p <= 0, 
\end{array}\right.
$$
where $p$ is the payoff of the previous round and the initial entrustment in the first round is three coins.

In the second stage, the number of entrusted coins is revealed, and both players decide whether to keep or return the coins entrusted to them. If the entrusted coins are returned, the other player receives double the amount of the entrustment. If the player instead decides to keep the coins, these coins are added to their coins.%

To analyze the effect of an emotional robot on trust and cooperation, the robot always returns the entrusted coins, except for round eight. 
The robot's cooperative strategy was chosen since research suggests that a strict strategy can hinder cooperation~\cite{Klapwijk2009} and that more generous game-play encourages cooperation~\cite{Weber2008}.

After round eight, trust and cooperation with the robot has to be re-established. The robot encourages regaining trust by acknowledging the trust violation, ``Perhaps I tried too hard to maximize my coins. I should not do that again.'' Hereby, the robot attempts to restore the broken trust. However, the statement is deliberately left open as promises have shown a strong effect on trust repair~\cite{Esterwood2021,Cominelli2021}.

To encourage cooperative game-play, the participants are unaware of the number of rounds to be played. Further, the participants are instructed to score the highest number of coins out of all participants, rather than competing against the robot.
As an incentive for participants, the number of obtained coins was anonymously placed on a public leader board.


\subsection{Experiment Setup}
\label{subsec:experiment_setup}

\begin{figure}[!t]
\vspace{0.3cm}
\centering
    \includegraphics[width=0.6\linewidth]{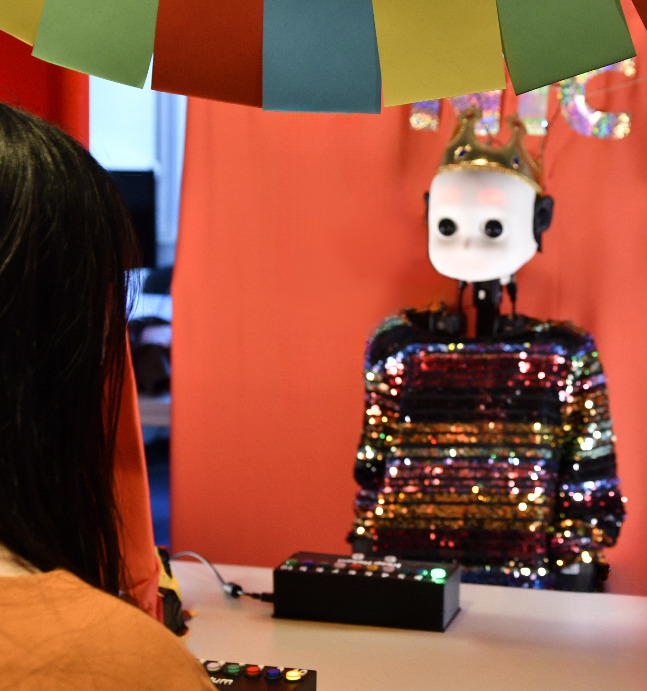}
    \caption{Interaction with the robot in the experiment.}
    \label{fig:nicos_exp}
\end{figure}

To avoid recognition of the study objective, the experiment is presented as a carnival attraction.
The general setup is illustrated in Figure~\ref{fig:nicos_stand}, and the interaction with the robot during the experiment is shown in Figure~\ref{fig:nicos_exp}.

\begin{figure}[h]
 \vspace{0.1cm}
\centering
    \includegraphics[width=0.9\linewidth]{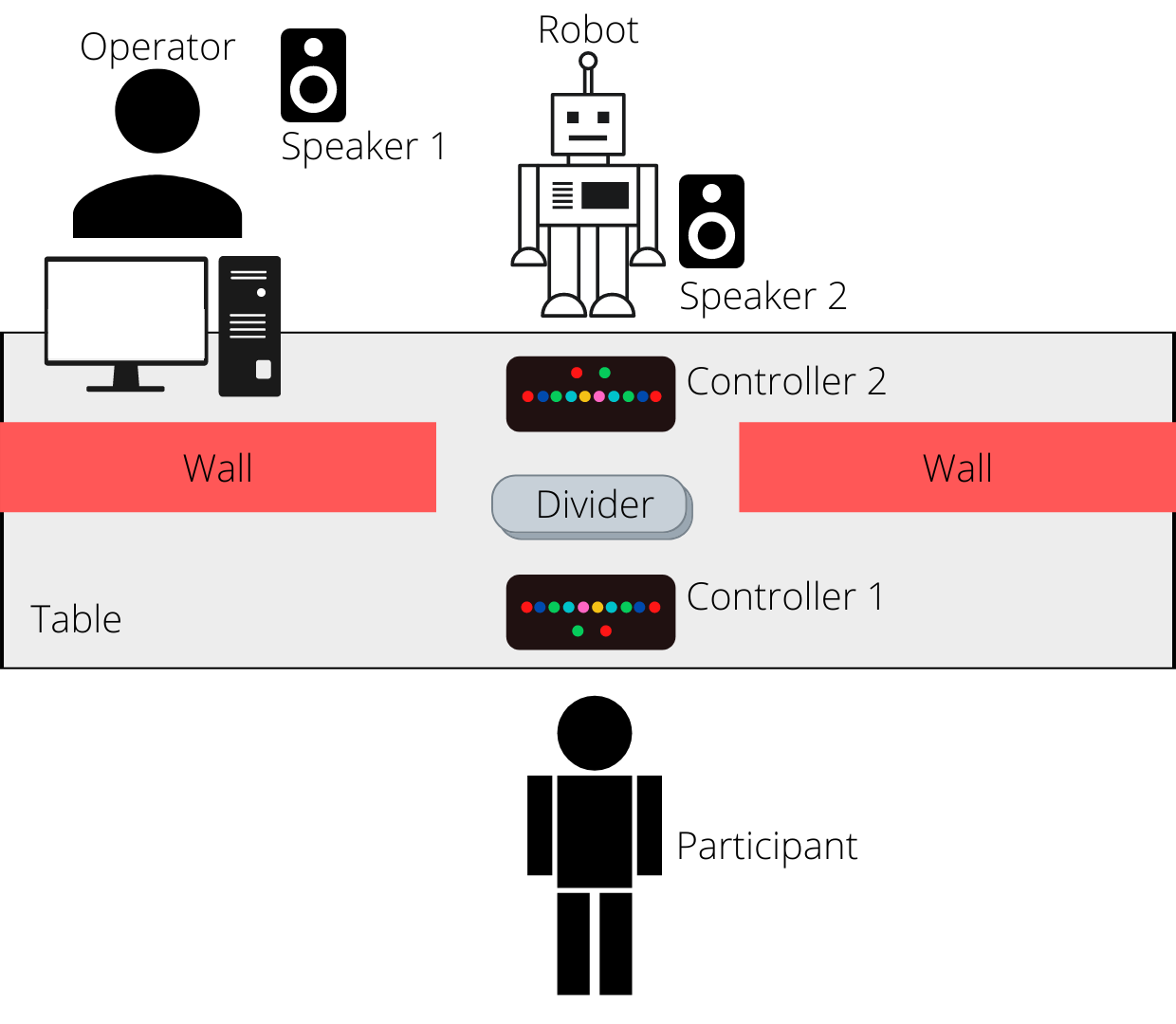}
    \caption{
        Schematic illustration of the experiment setup.
        The human operator is not visible to the participant.
    }
    \label{fig:schematic_setup}
\end{figure}

For a detailed illustration of the experiment setup, a schematic overview is provided in Figure~\ref{fig:schematic_setup}.
During the experiment, the participant is seated in front of a table and faces the robot. Both players (Participant and Robot) use a controller (Controller 1 and Controller 2) to input their decisions during the game. Each controller has two rows of light-up buttons. The first row allows the participant to select the number of coins to entrust by pressing one of the buttons labeled from 1 to 10. Similarly, with the second row of buttons, the participant can decide whether to keep or return the other player's entrustment. 
During both game stages, the robot utilizes a button-press animation, where the robot moves its hand close to the buttons and focuses its gaze on the controller.
Afterward, the lighted button of the robot's controller provides visual confirmation to the participant about the robot’s decision.




To block the players' view of the other player's controller during the decision-making of both game stages, a motorized divider is used. Additionally, the stages of each round are clearly separated by a narrator's voice that announces the decisions of each player.

The difference between the narrator and the robot is emphasized by utilising two spatially separated loudspeakers (Speaker 1 and Speaker 2).
The first loudspeaker is located close to the experiment operator and is used by the narrator for game-related information and announcements. The second loudspeaker is used for the robot's voice and is located inside the robot's body frame.

Before the experiment, a pilot study with eight participants was conducted. The data evaluation of the pilot study indicated that the participants understood the provided instructions and chose cooperative strategies instead of exploiting the robot's cooperative behavior. Further, it was suggested that the participants noticed the difference in emotions between both robots.

\subsection{Emotional Expressions}
\label{subsec:software_setup}


In the experiment, the Neuro-Inspired COmpanion (NICO)~\cite{kerzel2017nico} robot was used, which was designed to combine neurorobotics with human-robot interaction.
The software for the experiment is implemented in the Robot Operating System (ROS)~\cite{quigley2009ros} and is composed of multiple modules to control the robots' interactions during the game, speech, gestures, and facial expressions.

\subsubsection{Speech}
The speech generation module utilizes Google Text-to-Speech\footnote{\url{https://github.com/pndurette/gTTS}} (gTTS) to synthesize speech from text for the robot's voice. Spoken sentences comprise instructions for the beginning of the game, the game-play, and the final interaction. For the robot's responses after each game round, sentence variations are included to maintain the perceived robot's animacy. Depending on the participant's choice, 18 unique sentences for defect and 38 sentences for cooperation are implemented. 
To differentiate between the robots, the emotional robot uses interjections to express emotions~\cite{Tsiourti2019}.
When the participant returns the coins, the emotional robot might say: ``\emph{Hooray!} You gave the coins back'', while the non-emotional robot limits the answer to: ``You gave the coins back''. In contrast, when the participant keeps the coins, the answer ``I see you kept the coins'' for the non-emotional robot, is adjusted to ``\emph{Owww}, I see you kept the coins'' for the emotional robot.


The emotional interjections have specific prosody to convey either happy emotions when the participant cooperates, or sad emotions in the case of defect. The intonation of these interjections was simulated by applying a transfer learning text-to-speech model~\cite{jia2018transfer} to generate the utterances from audio references expressing either happiness or sadness. Afterward, the generated utterances were post-processed for artifact filtering, pitch, and tempo to match the robot's voice.
For the experiment, six different happiness-conveying interjections and five sadness-conveying interjections are utilized.

\subsubsection{Gestures and Facial Expressions}

The robot’s gestures are used to increase liveliness and display emotional behavior.
For the non-emotional robot, three neutral gestures (looking in a direction, pointing, and a hand gesture) are implemented~\cite{Tsiourti2019}. 
The happy and sad gestures of the emotional robot are designed to express emotions depending on the participant's decision to cooperate or defect.
For instance, gestures that include lowered arms and head are used to express sadness, while opening the arms is associated with happiness~\cite{de2004modeling}.
For each emotion, three different happy and sad gestures are implemented and randomly displayed after each round.





\begin{figure}[t]
 \vspace{0.3cm}
\centering
    \includegraphics[width=0.25\linewidth]{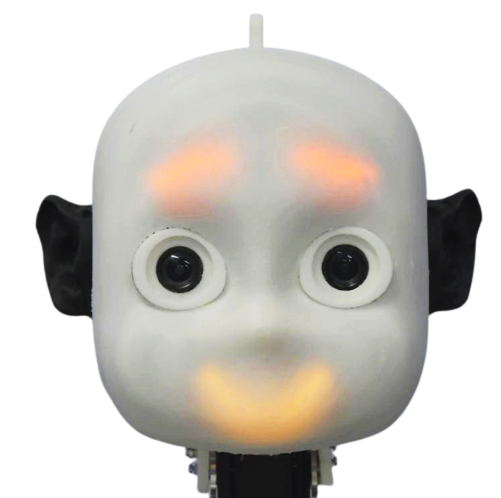}
    \includegraphics[width=0.25\linewidth]{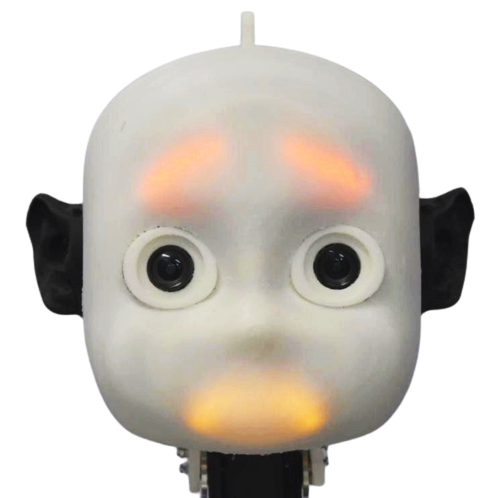}
    \includegraphics[width=0.25\linewidth]{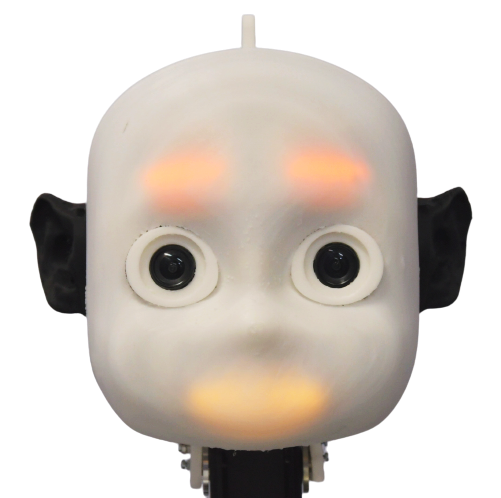}
    \caption{Facial expressions of the NICO robot. From left to right: happy, sad, neutral.}
    \label{fig:nico_face}
\end{figure}

In addition to the emotion-conveying gestures and interjections, the robot possesses LED arrays under its translucent face-plate in the mouth and eye area, which allows for displaying seven universal emotions~\cite{ekman1997universal}. Depending on the participant's decision to cooperate or defect, the robot shows either a happy or sad facial expression. During the remainder of the game and in the non-emotional group, a neutral facial expression is shown. The facial expressions of the NICO robot are depicted in Figure~\ref{fig:nico_face}.

\subsection{Questionnaires and Measurements}
\label{subsec:measurements}


For the evaluation of an emotional and non-emotional robot, the Godspeed~\cite{Bartneck2009} and Discrete Emotions~\cite{harmon2016discrete} questionnaires are assessed, and the participants' entrusted coins and decisions to cooperate or defect are recorded.

The Godspeed questionnaire provides standardized metrics about the perceived robot's anthropomorphism, animacy, likeability, intelligence, and safety.
Since the conducted experiment does not pose safety concerns, the category of perceived safety was omitted. 

The Discrete Emotions questionnaire provides insight into the participants' emotions during the experiment. Specifically, the questionnaire measures eight discrete emotions: anger, anxiety, desire, disgust, fear, happiness, relaxation, and sadness.

In each game round, the number of entrusted coins is a direct measure of the participant's trust in the robot.
Likewise, the decision to keep or return the robot's coins is a direct measure of cooperation.

\subsection{Study Design and Procedure} 
\label{subsec:measurements_study_design}

The experiment is conducted in a between-subject design with two groups. The robot in the emotional group shows happy and sad emotions through facial expressions, gestures, and utterances. These displays of emotion are shown at the end of each round. The robot in the non-emotional group also demonstrates facial expressions, speech, and gestures. However, the non-emotional robot's gestures are replaced with neutral gestures and a neutral facial expression is used. 



The participants signed a consent form agreeing to participate in the experiment, and were randomly assigned to one of the two experimental conditions. After the participant's demographics are assessed, the participant is handed a scenario description with an introduction to the experiment. Then, the participant is brought to a neighboring room where the experiment is conducted and seated in front of the robot. An introduction to the controller is provided, and a trial round introduces the experiment procedure. Afterward, the experiment begins. After the experiment, the participant is guided back to the initial room and the questionnaires are assessed.



\section{Results}
\label{sec:results}

\begin{figure}[!h]
\vspace{0.3cm}
    \centering
     \begin{subfigure}{0.435\textwidth}
        \includegraphics[width=1\linewidth]{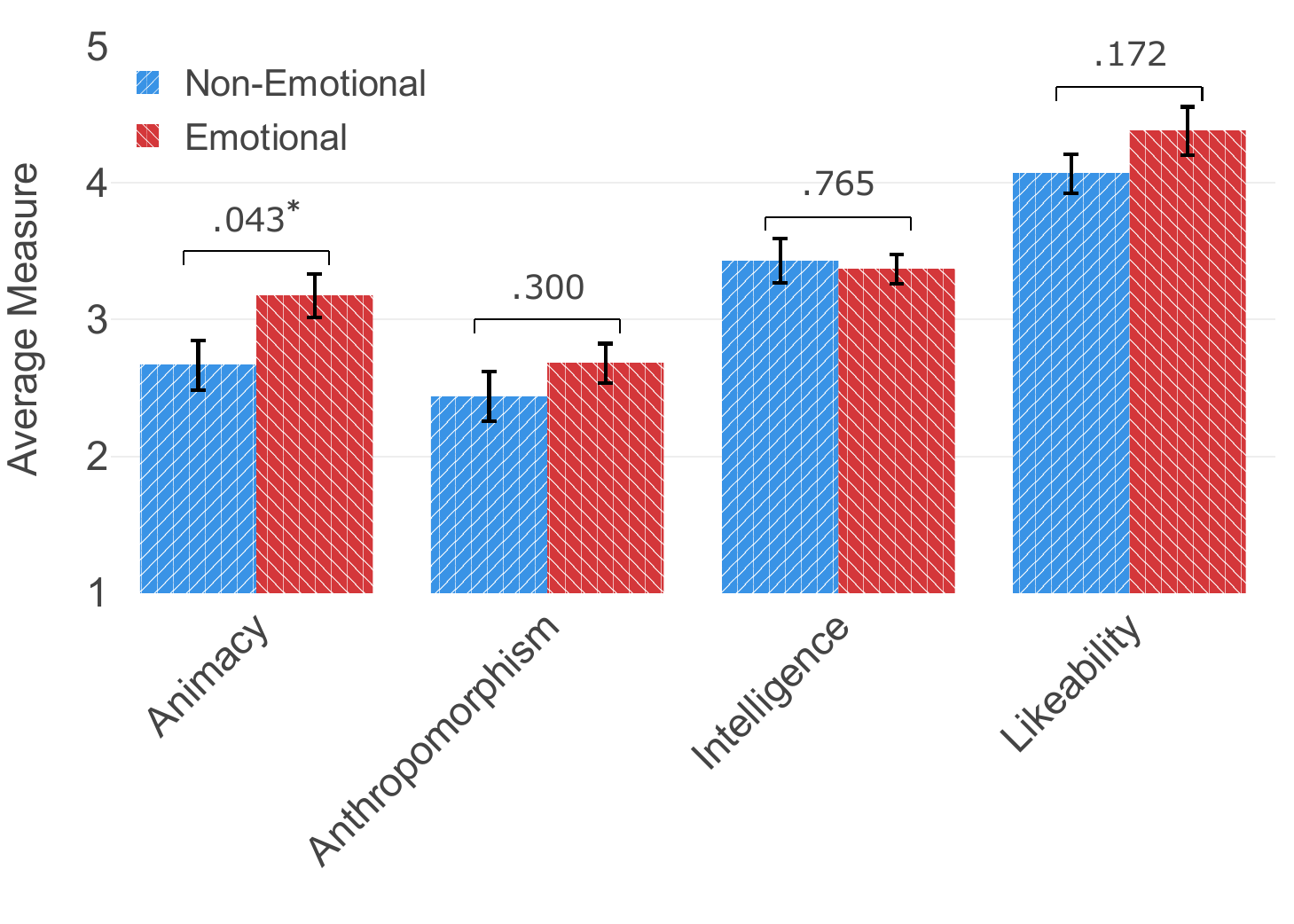}
    \caption{}
    \label{fig:1a}
  \end{subfigure}%
    

    
     \begin{subfigure}{0.435\textwidth}
        \includegraphics[width=1\linewidth]{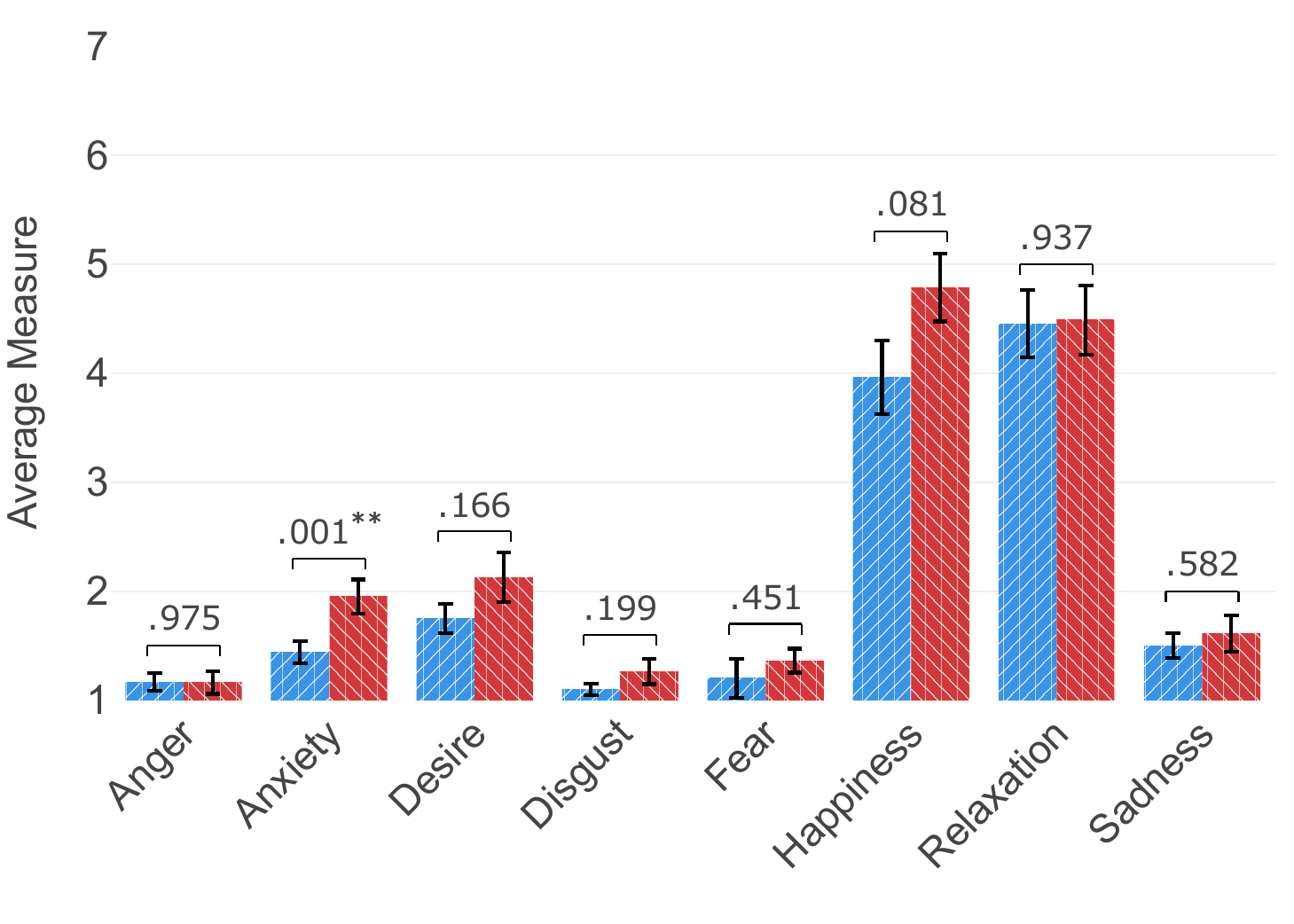}
    \caption{}
    \label{fig:1b}
    \end{subfigure}%

    \caption{Godspeed (a) and Discrete Emotions (b) items for the emotional and non-emotional group. Values indicate \textit{p} value, *\textit{p} $<$ .05, **\textit{p} $<$ .01 }
    \label{fig:godemo}
\end{figure}

An illustration of the Godspeed and Discrete Emotions questionnaire items with standard errors is shown in Figure~\ref{fig:godemo}.
A Student's \textit{t}-test~\cite{student1908probable} of the Godspeed questionnaire items shows a significant difference in animacy (\textit{p} = .043) of the emotional robot (\textit{M} = 3.18, \textit{SD} = 0.71) in contrast to the non-emotional robot (\textit{M} = 2.66, \textit{SD} = 0.84).
Regarding the participants' emotions, the Student's \textit{t}-test suggests a significant difference in the anxiety item (\textit{p} = .001) between the emotional group (\textit{M} = 1.95, \textit{SD} = 0.70) and non-emotional group (\textit{M} = 1.44, \textit{SD} = 0.46).

%



\begin{figure}[h!]
\vspace{0.3cm}
    \centering

\centering
     \begin{subfigure}{0.435\textwidth}
       \includegraphics[width=1.\linewidth]{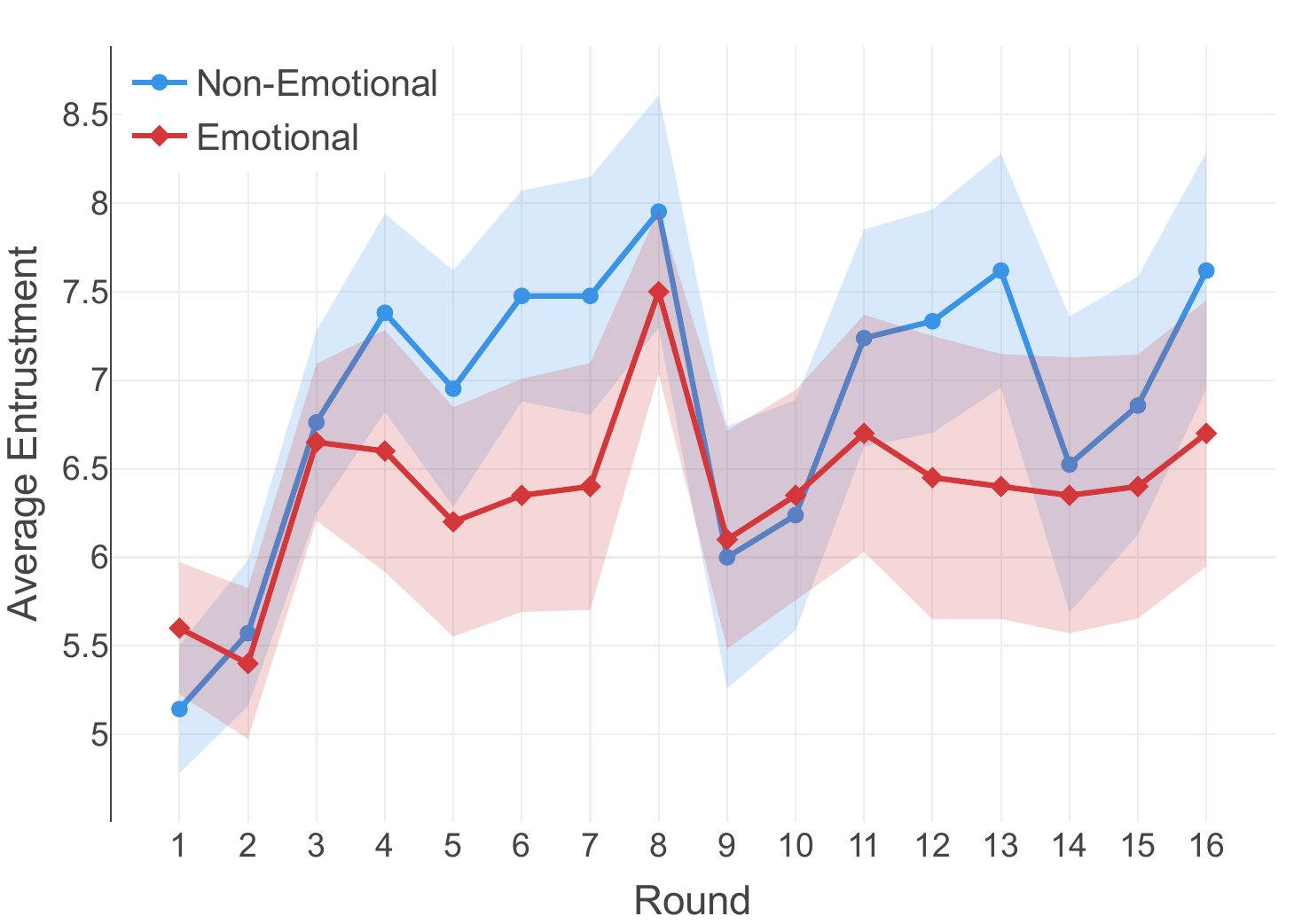}
    \caption{}
    \label{fig:2a}
  \end{subfigure}%
    
    \centering
     \begin{subfigure}{0.435\textwidth}
       \includegraphics[width=1.\linewidth]{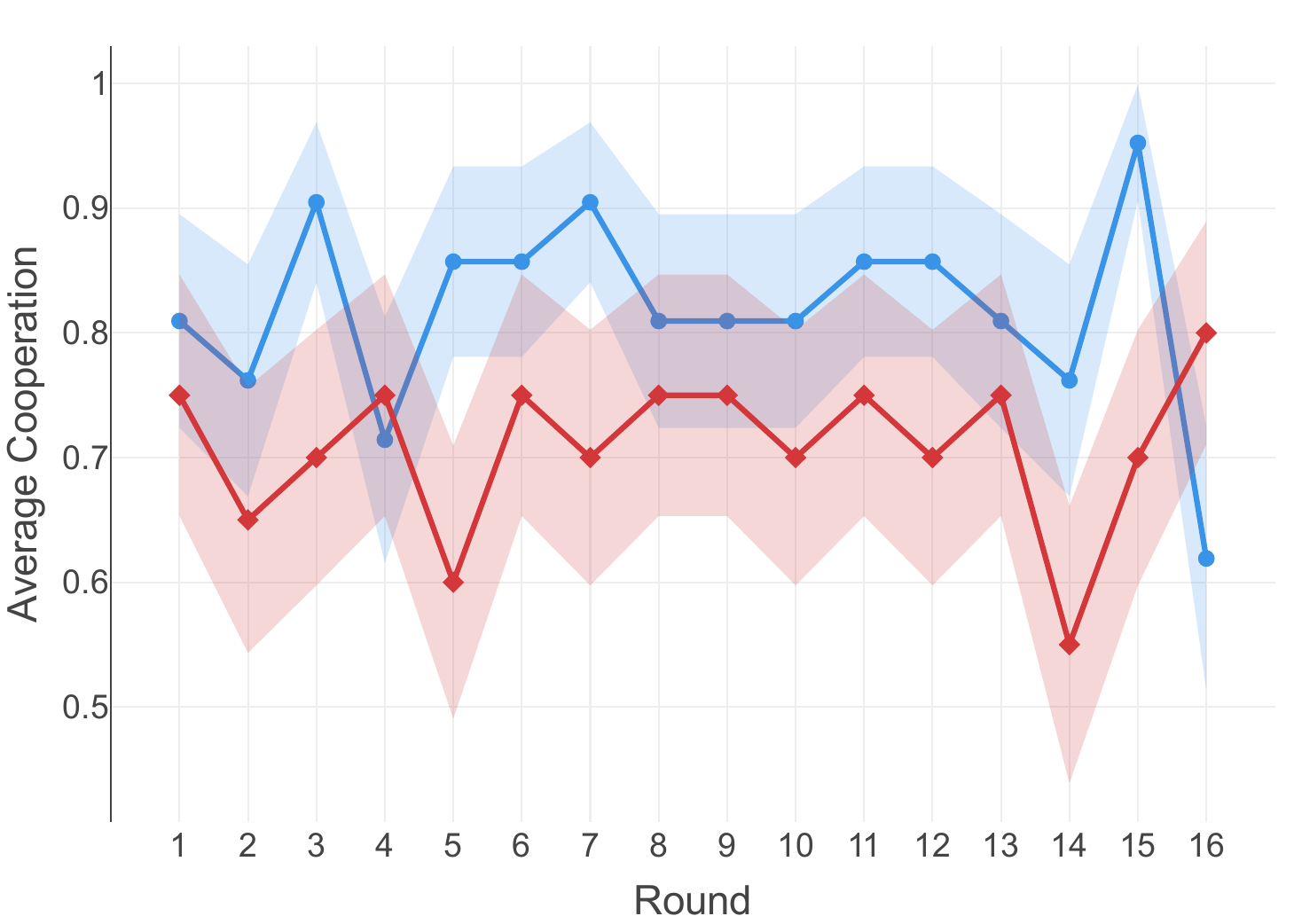}
    \caption{}
    \label{fig:2b}
  \end{subfigure}%

   \caption{Average entrustment (a) and cooperation rate (b) over the consecutive rounds.}
   \label{fig:coinscoop}
\end{figure}

Over the course of the experiment, the participants in both groups entrusted coins to the robot and decided to cooperate or defect. The on-average entrusted coins and cooperation rate with standard errors are shown in Figure~\ref{fig:coinscoop}.
A Mann–Whitney U test~\cite{mcknight2010mann} shows that the non-emotional group (\textit{M} = 6.88, \textit{SD} = 2.94) entrusted significantly more coins to the robot over the course of the experiment (\textit{p} = .015) than the emotional group (\textit{M} = 6.38, \textit{SD} = 2.85). 
For the cooperation rate, a Fisher's exact test~\cite{fisher1992statistical} shows that the non-emotional group (\textit{M} = 0.82, \textit{SD} = 0.39) was significantly more likely to cooperate (\textit{p} = .001) than the emotional group (\textit{M} = 0.71, \textit{SD} = 0.45).

Analyzing the difference of the average entrustment for both groups over two consecutive rounds, a Mann–Whitney U test suggests that the emotional group entrusted significantly more coins (\textit{p} = .038) in the third round (\textit{M} =  6.65, \textit{SD} = 1.98) in comparison to the second round (\textit{M} = 5.40, \textit{SD} = 1.90).
In the ninth round, after the robot defected, a noticeable decrease in the average entrustment can be noticed.
Especially, the average entrustment of the non-emotional group significantly (\textit{p} = .039) decreased from the eighth round (\textit{M} = 7.95, \textit{SD} = 3.01) to the ninth round (\textit{M} = 6.00, \textit{SD} = 3.39).


For the cooperation rate, a Fisher's exact test shows a significant difference (\textit{p} = .045) between the emotional group (\textit{M} = 0.70, \textit{SD} = 0.46) and non-emotional group (\textit{M} = 0.95, \textit{SD} = 0.21) in round 15, and
a difference in cooperation for the non-emotional group (\textit{p} = .020) between round 15 (\textit{M} = 0.95, \textit{SD} = 0.21) and round 16 (\textit{M} = 0.62, \textit{SD} = 0.49).
Although the robot's defect noticeably reduced the entrustment in the ninth round, the cooperation rate remained unaffected. Further, the participants exhibited a variety of entrustment strategies as indicated by the standard error.

To estimate the relationship between the assessed items of the Godspeed and Discrete Emotions questionnaire, the Spearman correlation~\cite{SpearmanRankCorrelation} of the on-average entrusted coins and cooperation rate is estimated and shown in Table~\ref{tab:ent_coop}.
The estimated correlations show a relationship between the entrusted coins and the perceived intelligence of the robot, where a robot that is perceived as more intelligent receives a larger entrustment. On the contrary, high positive anticipation as measured with the Desire item results in fewer entrusted coins and can decrease the likelihood of cooperating.
The amount of coins entrusted to the robot exhibits a strong correlation with the likelihood of cooperation.

\bgroup
\def\arraystretch{0.9}
\begin{table*}[h]
 \vspace{0.4cm}
\centering
\resizebox{\textwidth}{!}{\begin{tabular}{c|llll|llllllll|l}

   \multicolumn{14}{c}{\textbf{Entrustment}} \\
  \multicolumn{1}{c}{} &\multicolumn{4}{c}{Godspeed} & \multicolumn{8}{c}{Discrete Emotions}\\
  \cline{1-13}
Item & Animacy & Likeability & Anthropomorphism & Intelligence & Anger & Fear & Disgust & Sadness & Desire & Happiness & Relaxation &\multicolumn{2}{l}{ Anxiety} \\ 
       \cline{1-13}
  $\rho$  & -.210 & -.099 & -.229 & .332 & .168 & -.055 & -.118 & -.052 & -.352 & -.221 & .084 & \multicolumn{2}{l}{-.029}  \\ 
  \textit{p} value   & \phantom{-}.189 & \phantom{-}.536 & \phantom{-}.149 & \textbf{.034} & .294 & \phantom{-}.734 & \phantom{-}.463 & \phantom{-}.746 & \phantom{-}\textbf{.024} & \phantom{-}.166 & .600 & \multicolumn{2}{l}{\phantom{-}.858} \\ 
   \cline{1-13}

   \multicolumn{14}{c}{} \\

   \multicolumn{14}{c}{\textbf{Cooperation}} \\
   \multicolumn{1}{c}{} &\multicolumn{4}{c}{Godspeed} & \multicolumn{8}{c}{Discrete Emotions}\\
  \hline
Item& Animacy & Likeability & Anthropomorphism & Intelligence & Anger & Fear & Disgust & Sadness & Desire & Happiness & Relaxation & Anxiety & Entrustment \\ 
  \hline
  $\rho$ & -.228 & -.196 & -.155 & .150 & .222 & -.078 & -.217 & -.035 & -.418 & -.171 & .250 & -.129 & \phantom{$<$}.733 \\ 
  \textit{p} value    & \phantom{-}.151  & \phantom{-}.218  & \phantom{-}.332 & .349 & .163 & \phantom{-}.626 & \phantom{-}.173 & \phantom{-}.826 & \phantom{-}\textbf{.007} & \phantom{-}.284 & .115 & \phantom{-}.421 &  $<$\textbf{.001} \\ 
   \hline
\end{tabular}}
\caption{Correlation between the entrustment, the cooperation rate, and the measured questionnaire items.}
    \label{tab:ent_coop}
\end{table*}



\section{Discussion}
\label{sec:discussion}

The assessed Godspeed and Discrete Emotions questionnaires show significant differences between both experiment groups.
Although both robots utilize gestures, facial expressions, and voice, the emotional robot was perceived as more animated than the non-emotional robot. 
The robot's emotional reaction to the participant's choice to either cooperate or defect, might have led the participants to perceive the emotional robot as more lively and animated. 
Further, the participants in the emotional group experienced stronger emotions toward the emotional robot.
Specifically, participants in the emotional experiment group exhibited a greater level of anxiety than the non-emotional group.

Evidently, the display of human-like emotions by the robot caused the participants more discomfort in the researched scenario.
In the social dilemma that requires cooperation and reliance on the other player, participants' negative emotions reduced trust and cooperation. 
This indicates that the participants associated a more human-like behavior with a higher unpredictability of the robot's strategy, whereas a more neutral robot could be associated with pre-programmed and predictable actions to return the participant's entrustment. Likewise, the display of human emotions encouraged a more competitive game-play, as shown by the lower cooperation rate with the emotional robot.

The analysis of changes in trust and cooperation over two consecutive rounds shows that both experiment groups behave similarly and that the robot defecting results in a significant trust loss for the non-emotional group. However, the cooperation rate remains unaffected. This might be attributed to the robot's trust repair attempt by acknowledging the trust violation. Afterward, the robot continues to cooperate, which rebuilds the participants' trust. 

The correlation analysis of the entrusted coins and the measured questionnaire items shows that the participants are likely to entrust more coins to a robot that is perceived as intelligent. This suggests that an as intelligent perceived robot could positively influence the establishment of trust and willingness to cooperate.
The negative correlation between the Desire item of the Discrete Emotions questionnaire and the entrusted coins indicates that a participant's desire to succeed can lead to a more cautious entrustment. Since this task requires cooperation, a strong desire to succeed might result in smaller entrustments to assess the partner's willingness to cooperate. 
For the cooperation rate with the robot, the negative correlation of the desire to outcompete the other player could indicate that the participants initially establish cooperation with the robot, which results in larger entrustments from the robot, with the intent to keep the robot's coins if the amount appears profitable.
These results underline that emotions directly affect trust and cooperation in the coin entrustment game, and suggest that negative emotions towards the robot might hinder the formation of trust and cooperation. 

Contrary to the desire to accumulate coins by defecting the robot, the amount of coins entrusted to the robot exhibits a strong correlation with cooperation. Since the coins represent trust, it appears plausible that a larger entrustment reflects the participant's willingness to cooperate. However, the correlation is symmetrical and does not show that trust is independent of cooperation.
The willingness to cooperate could affect the extent of trust in the robot.


\section{Limitations}
\label{sec:limitations}



Despite the difference in trust and cooperation between an emotional and non-emotional robot, the findings have some limitations that should be addressed in future work.
The sample size of the experiment could be increased, which might lead to the identification of additional effects and emotions that influence trust and willingness to cooperate with a robot in a social dilemma.
Personality traits and trust disposition should be considered to estimate the effect and interplay between emotions and trust more accurately.
Additionally, the experiment was conducted in a laboratory environment, and the effects on trust and cooperation could be more pronounced in a real-world environment. Likewise, a more realistic-looking humanoid robot could have a stronger effect.



\section{Conclusion}
\label{sec:conclusion}

This study investigated the effects of an emotional robot on trust and cooperation in the coin entrustment game.
Specifically, a robot that conveys emotions by utilizing prosodic speech interjections, facial expressions, and body gestures was compared to a robot that portrays neutral emotions. The robot was programmed to cooperate throughout the experiment, except for the experiment's midpoint, to encourage the participants' trust and cooperation.
The participants' entrusted coins, cooperation, perception of the robot, and emotions during the experiment were evaluated.

The results show that the participants experienced more anxiety during the interaction with the emotional robot.
Accordingly, the emotional robot received less trust and participants were less likely to cooperate. The perceived robot's intelligence affected trust, and robots that exhibit emotions might be perceived as less suitable and competent for a cooperative task.  
However, emotional robots could be more resilient to breaches of trust, which points towards differences in interaction with and perception of emotional robots. These differences might be based on humans' preconceived assumptions about robots and how they should operate.
Furthermore, the results are consistent with findings in the literature that suggest that trust in anthropomorphic robots is task-dependent, and provide evidence that depending on the task, trust and cooperation with a neutral robot can be higher than with a robot that displays human-like emotions.

Subsequent experiments could research the specific effects of robots displaying either negative or positive emotions on trust and cooperation. Further, the granularity of the displayed emotions could be considered, to provide a better understanding of the impact of emotions in human-robot interaction.
Consequently, leading to guidelines and insights on applications for which emotional robots could be beneficial or have adverse effects.  
 

This research underlines the importance of social interaction between robots and humans, where an anthropomorphic robot can reduce trust. Especially, for critical or safety-related tasks, where cooperation is essential and mistakes can result in consequences, the relationship between emotions and the effect on trust and cooperation has to be considered.

	\bibliographystyle{ieeetr}
	\bibliography{bib.bib}
	
\end{document}